\documentclass{esannV2}
\usepackage{graphicx}
\usepackage{hyperref}
\usepackage[utf8]{inputenc}
\usepackage{amssymb,amsmath,array}
\DeclareMathOperator{\E}{\mathbb{E}}
\usepackage{setspace}

\begin{document}
\title{Capturing Variabilities \\from Computed Tomography Images \\with Generative Adversarial Networks}

\author{Umair Javaid$^{1,2}$ and John A. Lee$^{1,2}$
%
\thanks{Umair Javaid is funded by the F.N.R.S. T\'el\'evie grant no. 7.4625.16. John A. Lee is a Senior Research Associate with the F.N.R.S.}
%
\vspace{.3cm}\\
%
1- Universit\'e Catholique de Louvain - ICTEAM \\
Place du Levant 3 L5.03.02, 1348 Louvain-la-Neuve - Belgium 
%
\vspace{.1cm}\\
2- Universit\'e Catholique de Louvain - IREC/MIRO \\
Avenue Hippocrate 55 B1.54.07, 1200 Brussels - Belgium \\
}

\maketitle

\begin{abstract}
With the advent of Deep Learning (DL) techniques, especially Generative Adversarial Networks (GANs), data augmentation and generation are quickly evolving domains that have raised much interest recently. However, the DL techniques are data demanding and since, medical data is not easily accessible, they suffer from data insufficiency. To deal with this limitation, different data augmentation techniques are used. Here, we propose a novel unsupervised data-driven approach for data augmentation that can generate 2D Computed Tomography (CT) images using a simple GAN. The generated CT images have good global and local features of a real CT image and can be used to augment the training datasets for effective learning. In this proof-of-concept study, we show that our proposed solution using GANs is able to capture some of the global and local CT variabilities. Our network is able to generate visually realistic CT images and we aim to further enhance its output by scaling it to a higher resolution and potentially from 2D to 3D.  
\end{abstract}

\section{Introduction}
Deep learning algorithms are data demanding; meaning more data leads to more effectiveness. Medical image classification and segmentation tasks using deep learning often have limited data due to scarcity of annotated data. This is particularly true in the health care industry where access to medical data is strictly protected due to privacy concerns.
To bridge this limitation, data augmentation techniques can be used. 

A common practice for image data augmentation consists of geometric augmentations such as reflection, rotation, scaling and translation \cite{gonzalez2008morphological} of the input images. 
Other approaches are the use of medical phantoms \cite{court2010use}, \cite{chang2010development} in which medical data is simulated with a certain degree of realism. These approaches however, are
implemented manually so as to extend the training datasets. Therefore, there is a need of an automated data augmentation technique, which is generative in nature and offers an acceptable degree of realism. 

Recently, GANs  \cite{goodfellow2014generative} have raised much interest in the domain of data generation and are being used extensively for data augmentation. A standard GAN  consists of two convolutional neural networks namely, Discriminator \textit{D} and a Generator \textit{G}, which are trained to learn simultaneously. 
Following the minimax strategy, one network (\textit{D}) learns how to distinguish real samples from the ones generated by the generator where \textit{G} produces samples that can trick \textit{D} into believing they are real.


In this proof-of-concept study, we address the challenges of data augmentation by proposing an automatic pipeline that is generative and offers a certain degree of realism with the medical data. We explore the potential of GANs to capture the variabilities in CT images and generate identically looking CT images. The generated images can be used to augment the training datasets which automate the augmentation processing and potentially leverage the scarcity of medical data.  


The dataset examined in this study is the Lung CT images dataset publicly available at the Lung Image Database Consortium image collection (LIDC) \cite{armato2004lung}. The dataset consists of 267 Lung CT scans of dimensions $512 \times 512$. 

Our contribution in this study is two-fold. First, we propose an automated unsupervised data-driven approach using a simple GAN that can generate realistic CT images, which can be used to augment the training datasets for effective learning. To our knowledge, this is the first implementation for generating full CT image from a lower-resolution (in our case, $40 \times 40$ resolution). Second, using a simple GAN, we propose a robust framework that 
can be scaled to a higher resolution without changing any parameters of the model.

The remainder of this paper is organized as follows. Section 2 reviews the related work and outlines recent advancements in the domain of data generation using GANs. We present the proposed Methodology in Section 3. Experimental results are shown in Section 4, followed by Conclusion and Future Perspectives in Sections 5. 

\section{Related Work}
Goodfellow et al. \cite{goodfellow2014generative} proposed an adversarial approach to learn deep generative models namely, Generative Adversarial Networks (GANs). 
A number of GAN architectures have been proposed to enable stable training of generative models for producing realistic images. Recent progresses like Radford et al. \cite{radford2015unsupervised} have shown improved performances on synthesizing realistic images by using multiple strided and fractionally strided convolutional layers for both discriminator and generator networks. 
 D. Jiwoong et al. \cite{im2016generating} make use of recurrent neural networks to train a generative model in order to produce high-quality images.
    
	Other than designing different GAN architectures, several authors have tried to rectify GAN output by improving the training cycle with different criteria. For example, enhanced loss functions that aid in better convergence and learning of GANs. Moreover, Zhao et al. \cite{zhao2016energy} presented an energy-based GAN that offers optimal discriminator learning by minimizing an energy function where the energy is computed by an auto-encoder structured discriminator. Other researchers like Nowozin et al. \cite{nowozin2016f} and Chen et al. \cite{chen2016infogan} analyze GANs from an information theory perspective; learning representations that capture different latent factors to generate data samples. 
    
	GJ. Qi \cite{qi2017loss} has recently proposed a solution using an advanced loss function. Rather than having a fixed loss function, the paper shows that a loss function can be learnt and allows better separation between generated and real examples. Arjovsky et al. \cite{arjovsky2017wasserstein} proposed WGAN in which they use Wasserstein distance (a.k.a. Earth-Mover distance). It offers stable training of GANs but increases the training time. Bertholet et al. \cite{berthelot2017began} recently proposed Boundary Equilibrium GAN by proposing a new equilibrium enforcing method that is paired with a loss derived from the Wasserstein distance to train GAN.
    
	There are several studies that use GANs to produce medical data. For example, Nie et al. \cite{nie2017medical} use GANs to estimate CT images from MR images. Guibas et al. \cite{guibas2017synthetic} make use of a dual-stage GAN to generate retinal fundi images. Chuquicusma et al. \cite{chuquicusma2017fool} use GANs to generate realistic lung nodule samples. However, no study has been done for generating a full CT image and to our knowledge, this is the first attempt to generate full CT images from low-resolution input images.

    In this study, 
    we show that our simple GAN with a standard loss function is able to learn the CT image features and can generate almost realistic CT images from low-resolution input images.

\section{Methodology}

GANs are generative models that can learn mapping from a random noise vector \textit{z} to an output image. 
Generator \textit{G} takes as input a random noise vector \textit{z} (uniformly distributed) and outputs image $X_\mathrm{fake}=G(Z)$. Depending upon the feedback of \textit{D}, \textit{G} improves its output images until the \textit{D} cannot distinguish them from real images. Simultaneously, \textit{D} receives as input either the real (training image) or fake sample (generated image) and classifies the data produced by \textit{G} either as real or fake. The generator is initialized with a random noise vector while \textit{D} is trained with a small set of real images (training data). The objective of a GAN can be expressed as 
$\min_G$$\max_D$ $V(D,G)$ with 
\begin{eqnarray}
 V(D,G) = \E_{x \sim p_{data}(x)} [\log D(x)] + \E_{z \sim p_{z}(z)} [\log (1-D(G(z)))], 
\end{eqnarray}
where \textit{G} tries to minimize this objective against an adversarial \textit{D} that tries to maximize it. 
	

Our proposed GAN is a simple architecture with few number of layers in both the \textit{D} and \textit{G} networks that use the standard Cross Entropy loss function.

    \textit{D} consists of 4 convolutional layers and two dense layers whereas \textit{G} is a fully convolutional network - composed of 6 convolutional layers. Following \cite{radford2015unsupervised}, both networks use LeakyReLU as activation function except the last layer of \textit{D} that uses Sigmoid activation whereas, Linear activation is used at \textit{G} output. Dropout is used in \textit{D} and batch normalization is used in \textit{G}. \\

The detailed architecture of our network is as follows:\\

     \hspace*{2.5cm} \underline{\textbf{Discriminator}}\\
1. Conv with 256 filters and $3\times3$ kernel. LeakyReLU.\\
\hspace*{2.5cm} - Strided convolution with stride of 2.\\
\hspace*{2.5cm} - Dropout.\\
2. Conv with 128 filters and $3\times3$ kernel. LeakyReLU.\\
\hspace*{2.5cm} - Dropout.\\
3. Conv with 64 filters and $3\times3$ kernel. LeakyReLU.\\
	\hspace*{2.5cm}     - Dropout.\\
4. Conv with 32 filters and $3\times3$ kernel. LeakyReLU.\\
\hspace*{2.5cm}		- Dropout.\\
5. Fully connected layer with output dimension 128. LeakyReLU.\\
6. Fully connected layer with output dimension 1. Sigmoid activation.\\

      \hspace*{2.5cm} \underline{\textbf{Generator}}\\
1. Transposed Conv with 256 filters and $3\times3$ kernel. Stride = 2.\\ 
\hspace*{2.5cm}	- Batch Normalization. 
	 LeakyReLU.\\
2. Transposed Conv with 128 filters and $3\times3$ kernel. Stride = 2.\\
\hspace*{2.5cm}	     - Batch Normalization. LeakyReLU.\\
3. Transposed Conv with 64 filters and $3\times3$ kernel.\\ 
	\hspace*{2.5cm}     - Batch Normalization. LeakyReLU.\\
4. Transposed Conv with 32 filters and $3\times3$ kernel.\\ 
\hspace*{2.5cm}		- Batch Normalization. LeakyReLU.\\
5. Transposed Conv with 16 filters and $3\times3$ kernel.\\
	\hspace*{2.5cm}	- Batch Normalization. LeakyReLU.\\
6. Transposed Conv with 3 filters and $3\times3$ kernel. Linear Activation.

\section{Experiments and Results}

The input images were downsampled from $512\times512$ to $40\times40$. As a baseline analysis, only 32 CT images were used as input. Afterwards, to test the learning capability of our network, it was trained on 76 and 129 images. No pre-processing was applied to the training images besides intensity scaling to the range [0, 2]. 
	
    All networks were trained from the scratch using mini-batch stochastic gradient descent (SGD) with a mini-batch size of 16. All weights were initialized from a zero-centered Gaussian distribution with standard deviation 0.02. \textit{G} uses a noise vector to represent all the variations. The noise vector used was 100 dimensional, uniformly distributed in the range [-1, 1]. 
	
    In the LeakyReLU, the slope of the leak was set to 0.2 in both networks. RMSProp was used as an optimizer where the learning rate for discriminator was set to 0.0001 and for generator it was set to 0.0002. A small value of L2 norm (i.e. $10^{-5}$) was also used for both networks to avoid over-fitting. The value of batch normalization was set to 0.9 whereas a dropout of 0.6 was used in the discriminator network.
	
\begin{figure}[!h]
\vspace{-0.3cm}
\includegraphics[width=\textwidth]{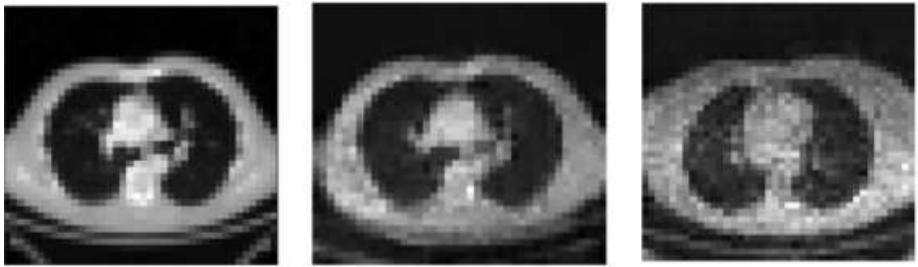}
\vspace{-0.85cm}\caption{Effect of kernel size on the generated images. Leftmost is the ground truth, middle is generated using $3\times3$ kernel size and the rightmost image is generated using $5\times5$ kernel size. 
}
\label{fig:1}
\end{figure}

\begin{figure}[!h]
\vspace{-0.6cm}
\includegraphics[width=\textwidth]{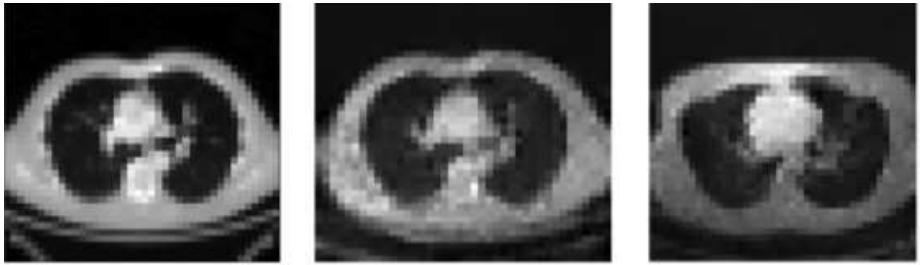}
\vspace{-0.8cm}\caption{Choice of optimizer influences the generated images. Leftmost is the ground truth, middle is generated using RMSProp and the rightmost image is generated using ADAM.}
\label{fig:2}
\end{figure}

\begin{figure}[!h]
\vspace{-0.5cm}
\includegraphics[width=\textwidth]{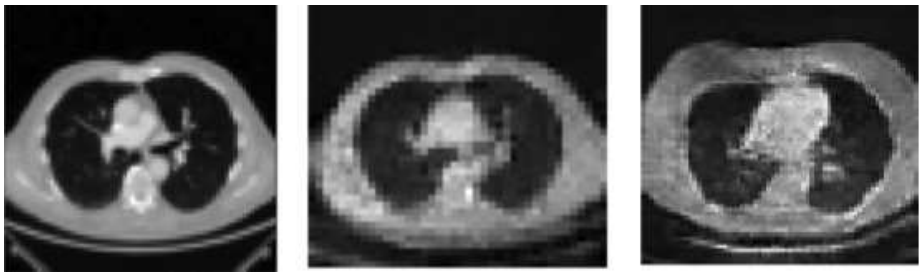}
\vspace{-0.7cm}\caption{Our proposed network is able to scale to a higher resolution without changing any parameter(s). Leftmost is the ground truth, middle is $40 \times 40$ and rightmost image is the generated $64\times64$ image.}
\label{fig:3}
\end{figure}

    Results from some of our experiments are given as follows. A comparison between the different convolutional kernel size is shown in Fig.~\ref{fig:1} 
followed by the effect of using RMSProp and ADAM as optimizers in Fig.~\ref{fig:2}. 
Fig.~\ref{fig:3} shows
the scalability of our network to a higher resolution i.e., from $40\times40$ to $64\times64$. 

We implemented our GAN in TensorFlow and for all the experiments, it was trained on a single NVIDIA GTX 1070 GPU. Training our model is not expensive, all the results shown were generated after 1 hour of training (30,000 epochs).

\section{Conclusion and Future Perspectives}
In this study, we proposed an automated augmentation framework using GANs. 
We showed that our approach is able to generate visually realistic CT images from low-resolution input images. 

As a proof-of-concept, we used a very simple GAN architecture that is able to capture CT variabilities.
We showed that our simple GAN offers good scalability to higher resolution images without changing any parameters. 



In future, we want to 
scale our network to higher resolutions, 
do a comparative study between a number of recent GAN architectures and an image quality analysis. As CT images are volumetric, we will also explore image generation in the 3D space which will retain the information lost in 2D space. Once higher resolution images are generated, we aim to use them in our training datasets for medical image segmentation tasks using Deep Learning.

--------------------------------------

\begin{footnotesize}
\begin{spacing}{0.5}
\bibliographystyle{unsrt}
\bibliography{main}
\end{spacing}
\end{footnotesize}


\end{document}